\documentclass[journal]{IEEEtran}

\ifCLASSINFOpdf
\else
\fi

\usepackage{epsfig}
\usepackage{graphicx}
\usepackage{amsmath}
\usepackage{amssymb}
\usepackage{algorithm}
\usepackage{algorithmicx}
\usepackage{algpseudocode}
\usepackage{subfig}
\usepackage{xcolor}

\def\eg{e.g. }
\graphicspath{{data/}}

\begin{document}

\title{Learning Hierarchical Features for Visual Object Tracking with Recursive Neural Networks}

\author{Li~Wang,~\IEEEmembership{Member,~IEEE},
		~Ting~Liu,~\IEEEmembership{Student Member,~IEEE},
		~Bing~Wang,~\IEEEmembership{Student Member,~IEEE},
		~Xulei~Yang,~\IEEEmembership{Member,~IEEE}
        and~Gang~Wang,~\IEEEmembership{Member,~IEEE}
\thanks{L. Wang and X. Yang are with Agency for Science, Technology and Research (A*STAR), Singapore. (e-mail: wa0002li@e.ntu.edu.sg; yangxl@i2r.a-star.edu.sg)}
\thanks{T. Liu, B. Wang and G. Wang are with Alibaba AI Labs, China. (e-mail: liut0016@e.ntu.edu.sg; wang0775@e.ntu.edu.sg; gangwang6@gmail.com)}}

{}

\maketitle

\begin{abstract}
Recently, deep learning has achieved very promising results in visual object tracking. Deep neural networks in existing tracking methods require a lot of training data to learn a large number of parameters. However, training data is not sufficient for visual object tracking as annotations of a target object are only available in the first frame of a test sequence. In this paper, we propose to learn hierarchical features for visual object tracking by using tree structure based Recursive Neural Networks (RNN), which have fewer parameters than other deep neural networks, \eg Convolutional Neural Networks (CNN). First, we learn RNN parameters to discriminate between the target object and background in the first frame of a test sequence. Tree structure over local patches of an exemplar region is randomly generated by using a bottom-up greedy search strategy. Given the learned RNN parameters, we create two dictionaries regarding target regions and corresponding local patches based on the learned hierarchical features from both top and leaf nodes of multiple random trees. In each of the subsequent frames, we conduct sparse dictionary coding on all candidates to select the best candidate as the new target location. In addition, we online update two dictionaries to handle appearance changes of target objects. Experimental results demonstrate that our feature learning algorithm can significantly improve tracking performance on benchmark datasets.
\end{abstract}


\IEEEpeerreviewmaketitle

\section{Introduction}

\IEEEPARstart{V}{isual} object tracking aims to locate a target object in a video sequence given its location in the first frame. It is a very challenging problem because target appearance may vary dramatically due to illumination change, partial occlusion, object deformation, etc. To solve these issues, some trackers \cite{DBLP:conf/cvpr/JiaLY12} employ local patch based appearance models to achieve very promising performance. 

Feature representation is very important to object tracking. Recently, deep learning based trackers \cite{DBLP:conf/nips/WangY13}\cite{DBLP:conf/bmvc/LiLP14}\cite{DBLP:journals/tip/WangLWCY15}\cite{DBLP:conf/icml/HongYKH15}\cite{DBLP:conf/iccv/WangOWL15}\cite{DBLP:conf/iccv/MaOHYY15} have achieved very promising performance by using learned hierarchical features rather than raw pixel values or hand-crafted features. Deep neural networks usually require a lot of training data to learn a large number of parameters. However, training data is not sufficient for visual object tracking as annotations of a target object are only available in the first frame of a test sequence.

To overcome this problem, existing feature learning based trackers pre-train their neural networks by using auxiliary data and then fine-tune network parameters according to specific target objects. Different from these methods, we propose to learn hierarchical features for visual object tracking by using tree structure based Recursive Neural Networks (RNN), which have fewer parameters than other deep neural networks, \eg Convolutional Neural Networks (CNN). As a result, our feature learning method does not require any network pre-training on auxiliary data and will not suffer from fine-tuning network parameters.

First, we learn RNN parameters to discriminate between target object and background in the first frame of a test sequence. Tree structure over local patches of an exemplar region is randomly generated by using a bottom-up greedy search strategy. Given the learned RNN parameters, we create two dictionaries regarding target regions and corresponding local patches based on the learned hierarchical features from both top and leaf nodes of multiple random trees. In each of the subsequent frames, we conduct sparse dictionary coding on all candidates to select the best candidate as the new target location. In addition, we online update two dictionaries to handle appearance changes of target objects.

The main contribution is that hierarchical features are learned to discriminate between target and background by using RNN which can successfully encode spatial information among local patches of a target object based on multiple random trees. RNN features learned at top nodes of random trees are able to capture structural information of target objects, which are robust to holistic appearance changes caused by illumination change or object deformation. Moreover, RNN features learned at leaf nodes represent local patches and capture local appearance changes due to partial occlusion. Therefore, our hierarchical features learned from both top and leaf nodes are beneficial for visual object tracking. Experimental results demonstrate that using our feature learning method can significantly improve tracking performance on the benchmark dataset \cite{DBLP:conf/cvpr/WuLY13}.

\section{Related Work}

\textbf{Visual Object Tracking.} During the past few decades, visual object tracking have received much attention. Many tracking algorithms have achieved very promising results. We refer interested readers to some recent surveys \cite{DBLP:conf/cvpr/WuLY13}\cite{DBLP:journals/pami/SmeuldersCCCDS14}. Our feature learning algorithm is integrated into a baseline tracker ASLA \cite{DBLP:conf/cvpr/JiaLY12} belonging to generative trackers which focus on modeling target appearance without considering background. Other generative trackers utilize subspace learning \cite{DBLP:journals/ijcv/BlackJ98}\cite{DBLP:journals/ijcv/RossLLY08}, sparse coding \cite{DBLP:conf/iccv/MeiL09}\cite{DBLP:conf/cvpr/LiSS11}\cite{DBLP:conf/cvpr/BaoWLJ12}, Gaussian mixture model \cite{DBLP:journals/ivc/McKennaRG99}, kernel-based model \cite{DBLP:journals/pami/ComaniciuRM03}, visual decomposition model \cite{DBLP:conf/cvpr/KwonL10}, etc. Although the baseline tracker in this paper is generative, our features are learned discriminatively to differentiate target objects from background. Discriminative trackers formulate object tracking as a binary classification problem. They use many machine learning algorithms such as SVM \cite{DBLP:journals/pami/Avidan04}\cite{DBLP:conf/iccv/TangBZT07}\cite{DBLP:conf/iccv/HareST11}\cite{DBLP:conf/cvpr/BaiT12}, boosting \cite{DBLP:conf/cvpr/GrabnerB06}\cite{DBLP:conf/iccv/LiuCL09}\cite{DBLP:conf/cvpr/ZeislLSB10}, graph embedding \cite{DBLP:conf/iccv/ZhangHML07}, multiple instance learning \cite{DBLP:conf/cvpr/BabenkoYB09}, metric learning \cite{DBLP:conf/eccv/WangHH10}\cite{DBLP:conf/cvpr/JiangLW12}, Gaussian process regression \cite{DBLP:conf/eccv/GaoLHX14}, etc.

\textbf{Deep Learning.} Recently, deep learning has achieved very promising results in visual object tracking \cite{DBLP:conf/nips/WangY13}\cite{DBLP:conf/bmvc/LiLP14}\cite{DBLP:journals/tip/WangLWCY15}\cite{DBLP:conf/icml/HongYKH15}\cite{DBLP:conf/iccv/WangOWL15}\cite{DBLP:conf/iccv/MaOHYY15}. Usually, deep neural networks require a lot of training data. However, only the first frame of a test sequence is annotated in visual object tracking. To overcome this problem, some deep learning based trackers pre-train their neural networks by using auxiliary data, \eg $80$ million tiny images dataset \cite{DBLP:conf/nips/WangY13}, face detection dataset \cite{DBLP:conf/bmvc/LiLP14} and Hans van Hateren natural scene videos \cite{DBLP:journals/tip/WangLWCY15}. In contrast, our feature learning method does not require any network pre-training on auxiliary data. Some other deep learning based trackers adopt existing deep neural networks, \eg R-CNN model built upon Caffe Library \cite{DBLP:conf/icml/HongYKH15} and VGG network pre-trained on ImageNet \cite{DBLP:conf/iccv/WangOWL15}, and then fine-tune network parameters based on training data of specific target objects. Different from these methods, we learn RNN parameters in the first frame of a test sequence and fix the parameters in the subsequent frames. Therefore, our method does not suffer from fine-tuning network parameters. 

\textbf{Recursive Neural Networks.} RNN has been successfully applied to natural language processing for sentiment analysis \cite{DBLP:conf/emnlp/SocherPHNM11}, phrase and sentence modeling \cite{DBLP:conf/emnlp/SocherHMN12} and paraphrase detection \cite{DBLP:conf/nips/SocherHPNM11}. Also, RNN is applied for parsing natural scene \cite{DBLP:conf/icml/SocherLNM11} and $3$D object classification \cite{DBLP:conf/nips/SocherHBMN12}. To our best knowledge, we are the first to learn hierarchical features by using RNN for visual object tracking.

\section{Learning Hierarchical Features Using RNN}

Deep learning has achieved very promising performance for visual object tracking. Some deep learning based trackers \cite{DBLP:conf/nips/WangY13}\cite{DBLP:conf/bmvc/LiLP14}\cite{DBLP:journals/tip/WangLWCY15} require auxiliary data to pre-train their neural networks. This kind of pre-training is necessary for feature learning but inconvenient for visual object tracking. Moreover, some other trackers \cite{DBLP:conf/icml/HongYKH15}\cite{DBLP:conf/iccv/WangOWL15} employ the deep neural networks pre-trained already and then fine-tune them during tracking. However, annotation of a target object in a test sequence is still limited for fine-tuning such deep neural networks with large numbers of parameters. To avoid inconvenient pre-training and fine-tuning, we propose to learn hierarchical features for visual object tracking by using tree-structure based RNN which has fewer parameters and hence requires less training data. We learn RNN parameters to discriminate target from background by using target annotation in the first frame and fix them for extracting hierarchical features from candidate regions in subsequent frames. Moreover, our learned hierarchical features are able to capture spatial information over local patches of a target region based on multiple random trees. This capability is beneficial for visual object tracking.

In this section, we present details of our feature learning algorithm. First, we give an overview of RNN. Then, we explain how to extract hierarchical features based on RNN. Next, we describe how to generate trees over local patches of target regions. Last, we depict how to discriminatively learn RNN parameters.

\begin{figure*}
	\begin{center}
		\includegraphics[width=1.0\linewidth]{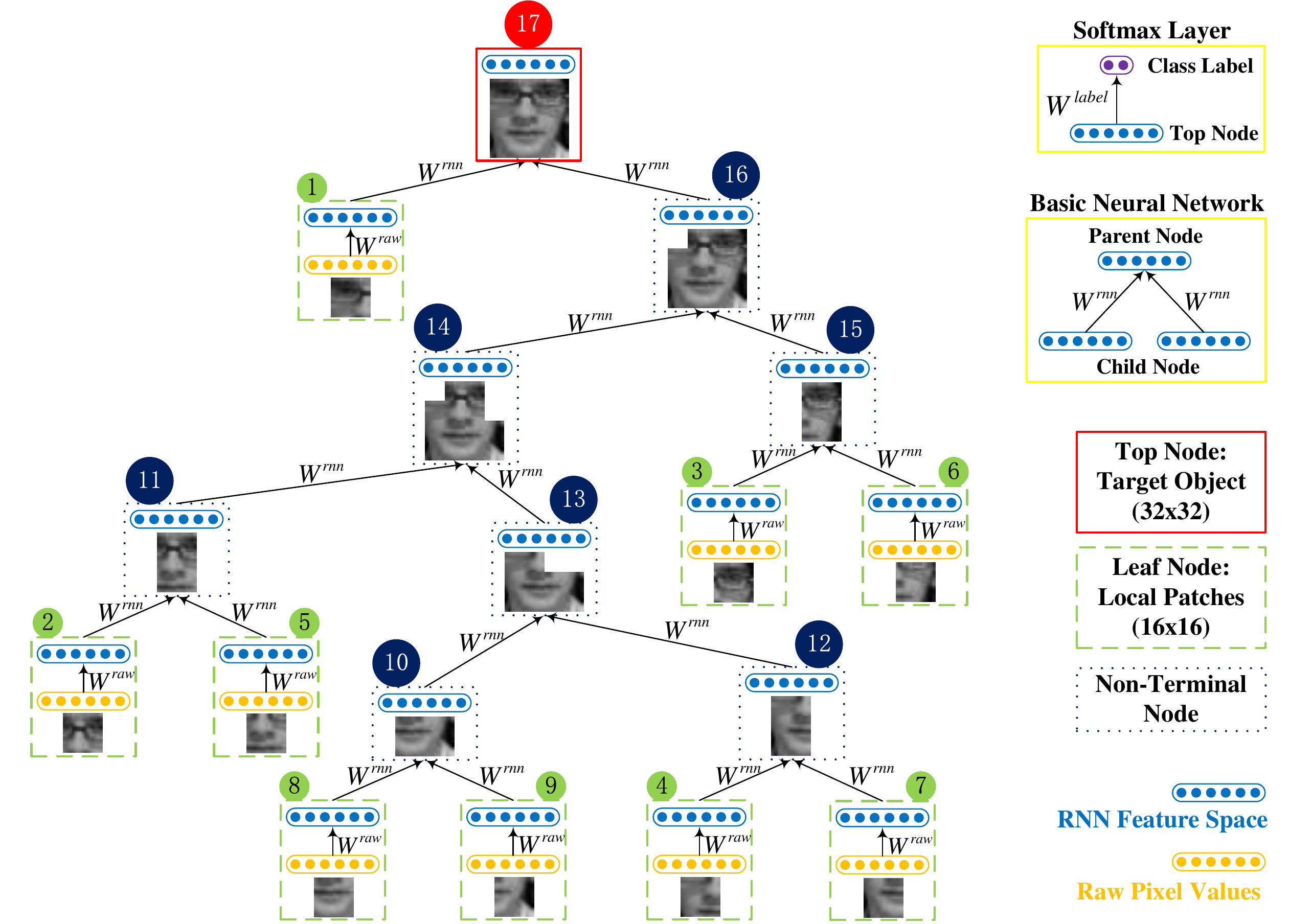}
	\end{center}
	\caption{Illustration of extracting hierarchical features using RNN with a known tree structure over the local patches of a target object.}
	\label{fig:extract_feat}
\end{figure*}

\subsection{Overview of RNN}
RNN is a deep neural network established by applying the same set of parameters recursively over a certain structure. In our case, RNN \cite{DBLP:conf/icml/SocherLNM11} is based on tree structure over local patches of a target object. There are three types of parameters: $W^{raw}$, $W^{rnn}$ and $W^{label}$. $W^{raw}$ and $W^{rnn}$ are used to extract hierarchical features from candidate regions based on multiple random trees. $W^{label}$ is to map extracted features regarding candidate regions into different classes (target object and background in this paper). In the first frame of a test sequence, we jointly learn these three types of parameters together to discriminate target from background. In the subsequent frames, we use the learned $W^{raw}$ and $W^{rnn}$ to extract hierarchical features for candidate regions.

\subsection{Extracting Features Using RNN}
Given a tree over local patches of a target region (see details in Section~\ref{sec:parse_tree}) and learned RNN parameters $W^{raw}$ and $W^{rnn}$ (see details in Section~\ref{sec:learn_param}), Figure~\ref{fig:extract_feat} illustrates an instance of extracting hierarchical features from a target region. We employ the local patch setting in ASLA \cite{DBLP:conf/cvpr/JiaLY12} and decompose a target object observation $x \in \mathbb{R}^{32\times32}$ into $9$ overlapping local patches $p \in \mathbb{R}^{16\times16}$. Each local patch can be vectorized into an $256$-dimensional raw pixel value feature $V_i \in \mathbb{R}^{256 \times 1}, i=1,\ldots,9$. Then, we utilize a neural network layer to map raw pixel values (orange circles in Figure~\ref{fig:extract_feat}) into a $n$-dimensional RNN feature space (blue circles). These RNN features at leaf nodes can be calculated as follows:
\begin{equation}
\zeta_i = f(W^{raw}V_i+b^{raw}),
\end{equation}
\noindent where $W^{raw} \in \mathbb{R}^{n\times256}$ is the transformation matrix, $b^{raw}$ is the bias, $f$ is the sigmoid function $f(x) = 1/(1+e^{-x})$ and $\zeta_i \in \mathbb{R}^{n \times 1}$ is the RNN feature at leaf nodes.

In the given tree, each node is associated with the same basic neural network illustrated in Figure~\ref{fig:extract_feat}. The basic network computes parent features based on two child input nodes as follows:
\begin{equation}
\eta_{(i,j)} = f([W^{rnn},W^{rnn}][\tau_i;\tau_j]+b^{rnn}),
\label{equ:parent_feat}
\end{equation}
\noindent where $W^{rnn} \in \mathbb{R}^{n \times n}$ is the transformation matrix, $b^{rnn}$ is the bias, $f$ is the sigmoid function $f(x) = 1/(1+e^{-x})$, $[\tau_i; \tau_j] \in \mathbb{R}^{2n \times 1}$ is the concatenated feature for two child nodes and $\eta_{(i,j)} \in \mathbb{R}^{n \times 1}$ is the parent feature. Note that the child node pair ($[\tau_i;\tau_j]$) possibly includes two leaf nodes ($[\zeta_i;\zeta_j]$), or two non-terminal nodes ($[\eta_i;\eta_j]$), or one leaf node and one non-terminal node ($[\zeta_i;\eta_j]$). Given the tree structure and the RNN parameters, we can extract hierarchical features from a target region at the top node of the RNN tree by recursively using the same basic neural network in a bottom-up manner. As a result, learned hierarchical features can capture spatial information over local patches of the target object. 

\begin{figure}
	\begin{center}
		\includegraphics[width=1.0\linewidth]{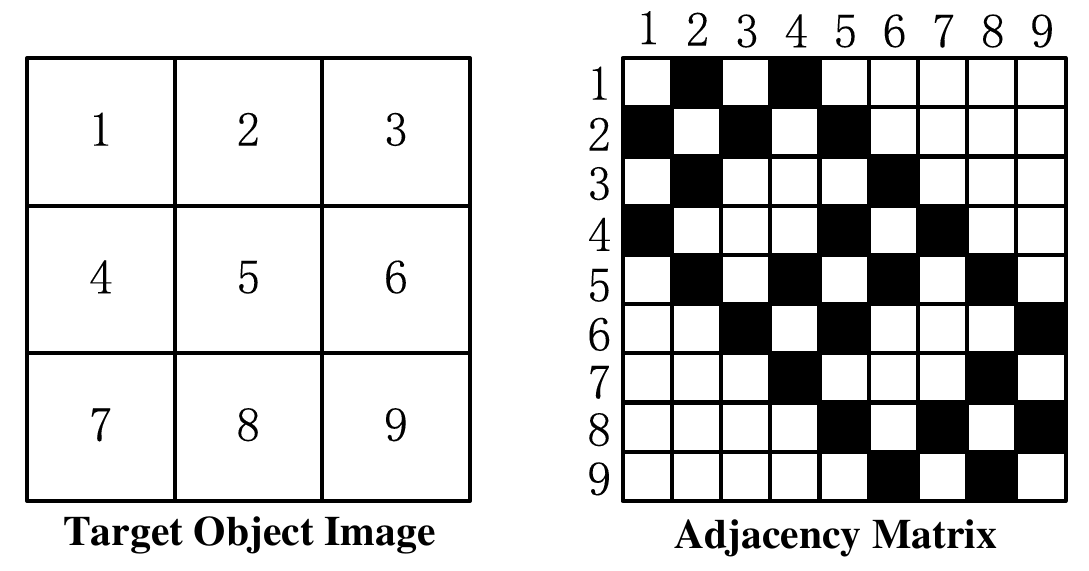}
	\end{center}
	\caption{Illustration of the spatial layout of local patches in a target object and the corresponding adjacency matrix. For example, patch $1$ and $2$ can be merged as they are spatially neighboring, so the corresponding matrix element is 1.}
	\label{fig:matrix}
\end{figure}

\subsection{Generating Random Trees}
\label{sec:parse_tree}
To extract hierarchical features using RNN, we need to generate a tree structure over local patches of a candidate region. There is no objective criterion to identify the best tree. Therefore, we randomly generate trees by using a greedy bottom-up searching strategy. We employ a number of trees together to extract RNN features, which are expected to be robust to certain noisy tree structure.

We first define an adjacency matrix $A$, where $A(i,j)=1$ if local patch $i$ and $j$ are spatially neighboring. It means that local patch $i$ and $j$ can be merged during generating random trees. Figure~\ref{fig:matrix} illustrates the spatial layout of local patches in a target object and the corresponding adjacency matrix, which is fixed in this paper.

Given the adjacency matrix $A$, we can find the pairs of neighboring local patches and denote the set of these potential child node pairs as follows:
\begin{equation}
C=\lbrace [p_i,p_j] | A(i,j)=1 \rbrace,
\end{equation}
\noindent where $p_i$ indicates the $i$th local patch. At the beginning (see Figure~\ref{fig:matrix}), we have the following pairs: $\lbrace$$[p_1,p_2]$, $[p_1,p_4]$, $\ldots$ , $[p_9,p_6]$, $[p_9,p_8]$$\rbrace$. Suppose we randomly select the pair $[p_i,p_j]$, we remove all pairs with either $p_i$ and $p_j$ from the set $C$ as follows:
\begin{equation}
\small
C = C - \lbrace [p_i, p_m] | m \in Neigh(i)\rbrace - \lbrace [p_j, p_n]| n\in Neigh(j) \rbrace,
\end{equation}
\noindent where $Neigh()$ denotes the neighborhood of a local patch. Then, we add new child pairs to the set $C$ as follows:
\begin{equation}
C = C \cup \lbrace [p_{(i,j)}, p_k] | k \in Neigh(i) \cup Neigh(j),~k \neq i,j \rbrace,
\end{equation}
\noindent where we randomly select another pair to merge. In the same way, we update the set $C$ and then repeat the previous steps until only one child node pair is left in the set $C$. Finally, we can achieve the top node of the tree over all local patches of a target region by merging the last child node pair in set $C$.    

We use the RNN tree shown in Figure~\ref{fig:extract_feat} as an example. Circles on top of nodes indicate the index numbers of nodes. We can observe that local patch $8$ and $9$ merge first and the corresponding child node pair $[p_8, p_9]$ is selected. Consequently, we update $C=\lbrace[p_1,p_2]$, $[p_1,p_4]$, $[p_2,p_1]$, $[p_2,p_3]$, $[p_2,p_5]$, $[p_3,p_2]$, $[p_3,p_6]$, $[p_4,p_1]$, $[p_4,p_5]$, $[p_4,p_7]$, $[p_5,p_2]$, $[p_5,p_4]$, $[p_5,p_6]$, $[p_6,p_3]$, $[p_6,p_5]$, $[p_7,p_4]$, $[p_{(8,9)}, p_5]$, $[p_{(8,9)}, p_6]$, $[p_{(8,9)}, p_7]$, $[p_5, p_{(8,9)}]$, $[p_6, p_{(8,9)}]$, $[p_7, p_{(8,9)}]\rbrace$. Next, we randomly select another pair in the updated $C$. In this example, local patch $2$ and $5$ are subsequently merged. Then, we repeat the previous manipulations until only one child node pair is left in the set $C$. As illustrated in Figure~\ref{fig:extract_feat}, we finally obtain the tree over local patches of the target region after merging local patch $1$ and non-terminal node $16$.
 
\subsection{Discriminative Learning of RNN Parameters}
\label{sec:learn_param}
Previously, deep learning based tracking methods require pre-training from auxiliary data or fine-tuning neural networks pre-trained already according to specific target object. In contrast, we learn RNN parameters only based on target annotation and background data in the first frame of a test sequence. Then, we fix the learned RNN parameters $W^{raw}$ and $W^{rnn}$ for extracting hierarchical features from candidate regions in the subsequent frames.

To discriminatively learn RNN parameters, we need to add a softmax layer with the parameter $W^{label}$, which connects RNN features at top nodes and class labels. Suppose we have $N$ training samples $\textbf{X} = \{x_1, \dots, x_N\}$ in terms of target and background, which are respectively sampled nearby and away from the target location in the first frame. We also have the corresponding labels $\textbf{L} = \{l_1^{GT}, \ldots, l_N^{GT}\}, l_i^{GT} \in \{[1;0], [0;1]\}$ ($[1;0]$ and $[0;1]$ indicate target and background respectively in our implementation). For each training sample $x_i$, RNN features can be extracted at the top node of a random tree based on the parameters $W^{raw}$ and $W^{rnn}$. Then, we apply the softmax layer illustrated in Figure~\ref{fig:extract_feat} for predicting class label:
\begin{equation}
l_i^{RNN}=softmax(W^{label}(\eta_{top}|T_i)),
\end{equation}
\noindent where $W^{label} \in \mathbb{R}^{2\times n}$ and $\eta_{top}$ is the learned RNN features at the top node of a random tree $T_i \in \mathcal{T}(x_i)$. $\mathcal{T}(x_i)$ is all possible trees over local patches of $x_i$. Consequently, we can compute the loss between the predicted label and the ground truth label for each training sample by using the cross-entropy error as follows:
\begin{equation}
\delta(x_i, l_i^{GT}|\theta, T_i) = -<l_i^{GT}, log(l_i^{RNN})>,
\end{equation}
\noindent where $\theta = \{W^{raw}, W^{rnn}, W^{label}\}.$ Finally, we have the objective function over all training samples as follows:
\begin{equation}
\Phi(\theta) = \frac{1}{N} \sum_{i=1}^N \delta(\theta) + \frac{\lambda}{2}\|\theta\|^2
\label{equ:obj}
\end{equation}
\noindent where $\lambda$ is the regularization parameter. The gradient of our objective function of Equation~\ref{equ:obj} w.r.t. the parameter set $\theta$ can be computed as follows:
\begin{equation}
\frac{\partial \Phi}{\partial \theta}=\frac{1}{N} \sum_i^N \frac{\partial \delta(\theta)}{\partial \theta} + \lambda \theta,
\end{equation}
\noindent The optimization can be performed by using backpropagation through structure \cite{goller1996learning}, which splits error messages at each node and then propagates to the child nodes. Then, we employ L-BFGS \cite{nocedal1980updating} to optimize our objective function.

\section{Our Tracking System}
Although the softmax layer of RNN can classify a candidate into target or background, it is relatively weak compared to the classifiers in state-of-the-art discriminative trackers. Therefore, we use the softmax layer to learn RNN parameters only and then integrate learned features into a state-of-the-art tracker ASLA \cite{DBLP:conf/cvpr/JiaLY12} with an adaptive local sparse appearance model. Interested readers may refer to ASLA \cite{DBLP:conf/cvpr/JiaLY12} for more details.

Suppose we have an observation set of target $x_{1:t}=\{x_1, \dots, x_t\}$ up to the $t^{th}$ frame and a corresponding feature representation set $z_{1:t}=\{z_1, \dots, z_t\}$, we can calculate the target state $y_t$ as follows:
\begin{equation}
y_t = \arg \max_{y_t^i} p\left( y_t^i | z_{1:t} \right ),
\end{equation}
\noindent where $y_t^i$ denotes the state of the $i^{th}$ sample in the $t^{th}$ frame. The posterior probability $p\left( y_t | z_{1:t} \right)$ can be inferred by the Bayes' theorem as follows:
\begin{equation}
p\left( y_t | z_{1:t} \right)\varpropto p\left( z_t | y_t \right)\int p\left( y_t | y_{t-1} \right) p\left( y_{t-1} | z_{1:t-1} \right) dy_{t-1}, \label{equ:postp}
\end{equation}
\noindent where $z_{1:t}$ denotes the feature representation, $p\left( y_t | y_{t-1} \right)$ denotes the motion model and $p\left( z_t | y_t \right)$ denotes the appearance model. In ASLA \cite{DBLP:conf/cvpr/JiaLY12}, the representations $z_{1:t}$ simply use raw pixel values of local patches, and the appearance model $p\left( z_t | y_t \right)$ employs sparse coding. In our tracker, we learn hierarchical features from raw pixel values by using RNN. First, we learn RNN parameters by using $20$ positive (nearby target) and $100$ negative (background) samples in the first frame of a test sequence. Then, we create two dictionaries regarding target regions and corresponding local patches based on learned RNN features respectively from top and leaf nodes of multiple random trees in the first $10$ frames. In each of subsequent frames, we conduct sparse coding on all candidate regions in terms of two dictionaries regarding target regions and local patches respectively. In addition, we online update two dictionaries every $5$ frames. In our implementations, we fix the dimension size of RNN feature space $n=50$, the regularization parameter $\lambda = 0.0001$ and the motion parameters at $[10,10,0.01,0,0.005,0]$ for all test sequences. We summarize our tracker using learned RNN features in Algorithm~\ref{alg:our_trac_sys}.

\begin{algorithm}[h]
\caption{\textbf{Our Tracker Using Learned RNN Features}} \label{alg:our_trac_sys}
\begin{algorithmic}[1]

\State \textbf{Input:} the last target state $y_{t-1}$, the RNN parameters $\theta$ learned in the first frame and two existing dictionaries respectively regarding target regions and local patches.

\State Generate $600$ target state candidates $y_{t}^i$ and the corresponding image observations $x_t^i$ nearby the previous target state $y_{t-1}$.

\State Extract hierarchical features $z_t^i$ from each image observation $x_t^i$ by using the RNN parameter $\theta$ learned in the first frame.

\State Calculate the posterior probability $p\left( y_t^i | z_{1:t} \right)$ according to Equation~\ref{equ:postp}.

\State Predict the target state by $y_t = \arg \max_{y_t^i} p\left( y_t^i | z_{1:t} \right )$.

\State Update two dictionaries every $5$ frames with learned RNN features respectively from top and leaf nodes of multiple random trees over the target region regarding the predicted state $y_{t}$.

\State \textbf{Output:} the predicted target state $y_{t}$ and two updated dictionaries.

\end{algorithmic}
\end{algorithm}

\section{Discussion}
\textbf{Parameter Size.} As sharing parameters in different layers of tree structures, RNN has fewer parameters than other deep neural networks, \eg Convolutional Neural Networks (CNN). The parameter set for RNN is $\theta = \{W^{raw}, W^{rnn}, W^{label}\}$, where $W^{raw} \in \mathbb{R}^{50\times 256}$, $W^{rnn} \in \mathbb{R}^{50 \times 50}$ and $W^{label} \in \mathbb{R}^{2\times 50}$. We can find that our RNN model has totally $15400$ parameters, which are much less than CNN (\eg 60 million parameters in \cite{DBLP:conf/nips/KrizhevskySH12}). Therefore, our feature learning algorithm using RNN does not require any pre-training and fine-tuning, which are actually not easy for visual tracking applications. 

\textbf{Computational Cost.} We run experiments on a PC without using any multi-core setting. The baseline tracker ASLA \cite{DBLP:conf/cvpr/JiaLY12} can achieve about $3$ fps. Our tracker using learned hierarchical features can achieve about $1.5$ fps. We have two major factors affecting efficiencies of our tracker: i) Learning RNN parameters in the first frame of a test sequence; ii) Extracting hierarchical features from candidate regions based on multiple random trees in each frame of the subsequent frames. As extracting features from $600$ candidates consumes too much time, we use a coarse-to-fine strategy which first identifies $20$ promising candidates according to tracking results from the baseline tracker ASLA \cite{DBLP:conf/cvpr/JiaLY12} and then ranks these $20$ candidates based on our learned hierarchical features. Anyway, the main objective in this paper is to show that our learned hierarchical features can improve tracking performance. The efficiency of our tracker could be enhanced by using parallel programming skills (\eg parfor in MATLAB) or other advanced hardware (\eg GPU).

\begin{figure*}
	\begin{center}
		\includegraphics[width=0.497\linewidth]{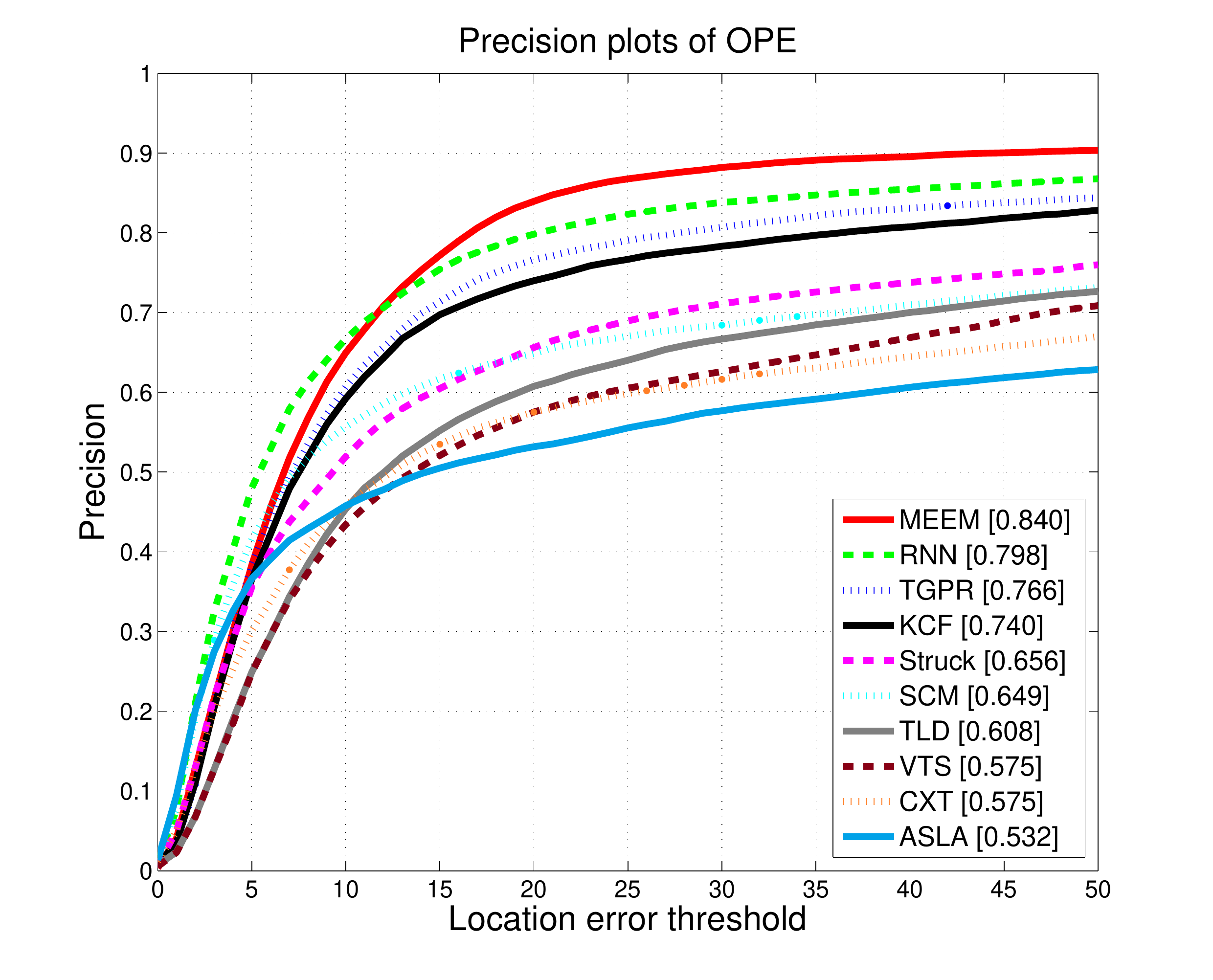}
		\includegraphics[width=0.497\linewidth]{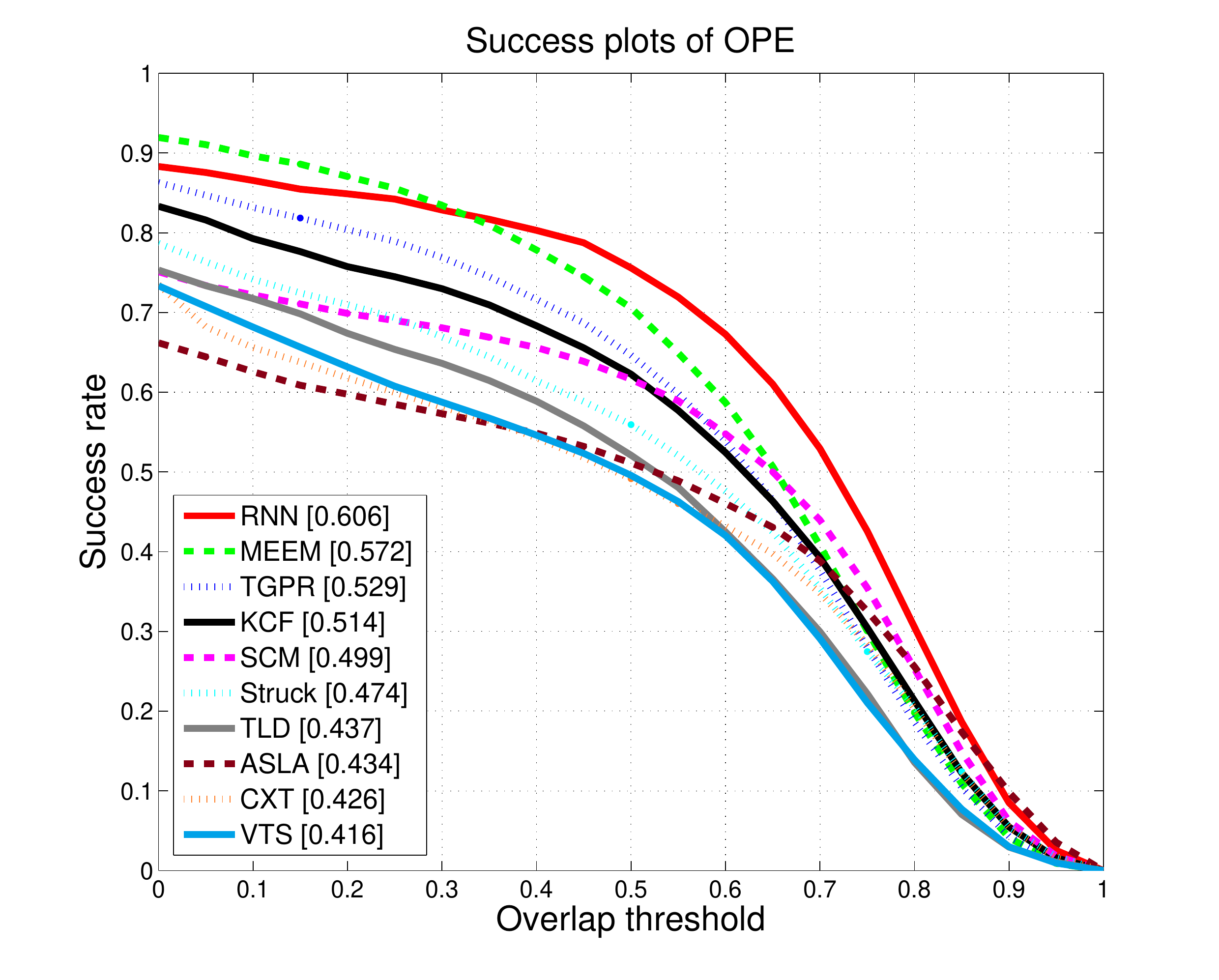}	
	\end{center}
	\caption{Precision and success plots of the tracking results from the top $10$ trackers on all sequences of the benchmark \cite{DBLP:conf/cvpr/WuLY13}.}
	\label{fig:pre+suc_plot_all}
\end{figure*}

\begin{table*}
\caption{Average precision scores on different attributes: background clutter (BC), deformation (DEF), fast motion (FM), in-plane rotation (IPR), illumination variation (IV), low resolution (LR), motion blur (MB), occlusion (OCC), out-of-plane rotation (OPR), out-of-view (OV) and scale variation (SV). The best and the second best results are in red and blue respectively.} 
\begin{center}
\begin{tabular}{|c||c|c|c|c|c|c|c|c|c|c|}
\hline
&\scriptsize{\textbf{ASLA}\cite{DBLP:conf/cvpr/JiaLY12}}&\scriptsize{\textbf{CXT}\cite{DBLP:conf/cvpr/DinhVM11}}&\scriptsize{\textbf{KCF}\cite{DBLP:journals/pami/HenriquesC0B15}}&\scriptsize{\textbf{MEEM}\cite{DBLP:conf/eccv/ZhangMS14}}&\scriptsize{\textbf{SCM}\cite{DBLP:conf/cvpr/ZhongLY12}}&\scriptsize{\textbf{Struck}\cite{DBLP:conf/iccv/HareST11}}&\scriptsize{\textbf{TGPR}\cite{DBLP:conf/eccv/GaoLHX14}}&\scriptsize{\textbf{TLD}\cite{DBLP:conf/cvpr/KalalMM10}}&\scriptsize{\textbf{VTS}\cite{DBLP:conf/iccv/KwonL11}}&\scriptsize{\textbf{RNN}}\\
\hline \hline
\scriptsize{\textbf{BC}}&0.496&0.443&0.753&\textcolor{red}{0.808}&0.578&0.585&0.761&0.428&0.578&\textcolor{blue}{0.779}\\
\hline
\scriptsize{\textbf{DEF}}&0.445&0.422&0.740&\textcolor{red}{0.859}&0.586&0.521&\textcolor{blue}{0.768}&0.512&0.487&0.762\\
\hline
\scriptsize{\textbf{FM}}&0.253&0.515&0.602&\textcolor{red}{0.757}&0.333&0.604&0.575&0.551&0.353&\textcolor{blue}{0.624}\\
\hline
\scriptsize{\textbf{IPR}}&0.511&0.610&0.725&\textcolor{red}{0.809}&0.597&0.617&0.706&0.584&0.579&\textcolor{blue}{0.766}\\
\hline
\scriptsize{\textbf{IV}}&0.517&0.501&\textcolor{blue}{0.728}&\textcolor{red}{0.778}&0.594&0.558&0.687&0.537&0.573&0.726\\
\hline
\scriptsize{\textbf{LR}}&0.156&0.371&0.381&0.494&0.305&\textcolor{blue}{0.545}&0.539&0.349&0.187&\textcolor{red}{0.660}\\
\hline
\scriptsize{\textbf{MB}}&0.278&0.509&\textcolor{blue}{0.650}&\textcolor{red}{0.740}&0.339&0.551&0.578&0.518&0.375&0.623\\
\hline
\scriptsize{\textbf{OCC}}&0.460&0.491&\textcolor{red}{0.749}&\textcolor{blue}{0.814}&0.640&0.564&0.708&0.563&0.534&0.743\\
\hline
\scriptsize{\textbf{OPR}}&0.518&0.574&0.729&\textcolor{red}{0.853}&0.618&0.597&0.741&0.596&0.604&\textcolor{blue}{0.780}\\
\hline
\scriptsize{\textbf{OV}}&0.333&0.510&\textcolor{blue}{0.650}&\textcolor{red}{0.730}&0.429&0.539&0.495&0.576&0.455&0.556\\
\hline
\scriptsize{\textbf{SV}}&0.552&0.550&0.679&\textcolor{red}{0.808}&0.672&0.639&0.703&0.606&0.582&\textcolor{blue}{0.750}\\
\hline \hline
\scriptsize{\textbf{Average}}&0.532&0.575&0.740&\textcolor{red}{0.840}&0.649&0.656&0.766&0.608&0.575&\textcolor{blue}{0.798}\\
\hline
\end{tabular}
\end{center}
\label{tab:pre_attr}
\end{table*}

\section{Experiments}
\textbf{Benchmark.} We evaluate tracking performance on a recent public benchmark \cite{DBLP:conf/cvpr/WuLY13} containing $50$ sequences which covers almost all challenging issues such as illumination changes, pose variations, occlusion, in/out-of-plane motions and cluttered background. The benchmark dataset reports the results from $29$ trackers. Here, we compare our tracker ``RNN" with the top $6$ trackers in the benchmark: ASLA \cite{DBLP:conf/cvpr/JiaLY12}, CXT \cite{DBLP:conf/cvpr/DinhVM11},  SCM \cite{DBLP:journals/tip/ZhongLY14}, Struck \cite{DBLP:conf/iccv/HareST11}, TLD \cite{DBLP:conf/cvpr/KalalMM10} and VTS \cite{DBLP:conf/iccv/KwonL11}. In addition, we also compare our tracker with $3$ recent state-of-the-art trackers: KCF \cite{DBLP:journals/pami/HenriquesC0B15}, MEEM \cite{DBLP:conf/eccv/ZhangMS14} and TGPR \cite{DBLP:conf/eccv/GaoLHX14}, which have reported their results on the benchmark.

The benchmark uses two measurements: i) Precision vs. Location error threshold, the percentage of the frames in which distances between tracking results and ground truth are below certain thresholds. ii) Success rate vs. Overlap threshold, the percentage of the frames in which overlapping percentages of tracking results against ground truth are higher than certain thresholds. We rank different trackers according to location error thresholding at 20 pixels for precision and Area Under Curve (AUC) for success rate. In addition, we use the one-pass evaluation (OPE) setting in the benchmark.

\begin{table*}
\caption{Average success rates on different attributes: background clutter (BC), deformation (DEF), fast motion (FM), in-plane rotation (IPR), illumination variation (IV), low resolution (LR), motion blur (MB), occlusion (OCC), out-of-plane rotation (OPR), out-of-view (OV) and scale variation (SV). The best and the second best results are in red and blue respectively.} 
\begin{center}
\begin{tabular}{|c||c|c|c|c|c|c|c|c|c|c|}
\hline
&\scriptsize{\textbf{ASLA}\cite{DBLP:conf/cvpr/JiaLY12}}&\scriptsize{\textbf{CXT}\cite{DBLP:conf/cvpr/DinhVM11}}&\scriptsize{\textbf{KCF}\cite{DBLP:journals/pami/HenriquesC0B15}}&\scriptsize{\textbf{MEEM}\cite{DBLP:conf/eccv/ZhangMS14}}&\scriptsize{\textbf{SCM}\cite{DBLP:conf/cvpr/ZhongLY12}}&\scriptsize{\textbf{Struck}\cite{DBLP:conf/iccv/HareST11}}&\scriptsize{\textbf{TGPR}\cite{DBLP:conf/eccv/GaoLHX14}}&\scriptsize{\textbf{TLD}\cite{DBLP:conf/cvpr/KalalMM10}}&\scriptsize{\textbf{VTS}\cite{DBLP:conf/iccv/KwonL11}}&\scriptsize{\textbf{RNN}}\\
\hline \hline
\scriptsize{\textbf{BC}}&0.408&0.338&0.535&\textcolor{blue}{0.578}&0.450&0.458&0.543&0.345&0.428&\textcolor{red}{0.587}\\
\hline
\scriptsize{\textbf{DEF}}&0.372&0.324&0.534&\textcolor{red}{0.582}&0.448&0.393&0.556&0.378&0.368&\textcolor{blue}{0.570}\\
\hline
\scriptsize{\textbf{FM}}&0.247&0.388&0.459&\textcolor{red}{0.568}&0.296&0.462&0.441&0.417&0.300&\textcolor{blue}{0.482}\\
\hline
\scriptsize{\textbf{IPR}}&0.425&0.452&0.497&\textcolor{blue}{0.535}&0.458&0.444&0.487&0.416&0.416&\textcolor{red}{0.576}\\
\hline
\scriptsize{\textbf{IV}}&0.429&0.368&0.493&\textcolor{blue}{0.548}&0.473&0.428&0.486&0.399&0.429&\textcolor{red}{0.563}\\
\hline
\scriptsize{\textbf{LR}}&0.157&0.312&0.312&0.367&0.279&\textcolor{blue}{0.372}&0.351&0.309&0.168&\textcolor{red}{0.520}\\
\hline
\scriptsize{\textbf{MB}}&0.258&0.369&\textcolor{blue}{0.497}&\textcolor{red}{0.565}&0.298&0.433&0.440&0.404&0.304&0.485\\
\hline
\scriptsize{\textbf{OCC}}&0.376&0.372&0.514&\textcolor{blue}{0.563}&0.487&0.413&0.494&0.402&0.398&\textcolor{red}{0.565}\\
\hline
\scriptsize{\textbf{OPR}}&0.422&0.418&0.495&\textcolor{blue}{0.569}&0.470&0.432&0.507&0.420&0.425&\textcolor{red}{0.579}\\
\hline
\scriptsize{\textbf{OV}}&0.312&0.427&\textcolor{blue}{0.550}&\textcolor{red}{0.597}&0.361&0.459&0.431&0.457&0.443&0.477\\
\hline
\scriptsize{\textbf{SV}}&0.452&0.389&0.427&0.510&\textcolor{blue}{0.518}&0.425&0.443&0.421&0.400&\textcolor{red}{0.573}\\
\hline \hline
\scriptsize{\textbf{Average}}&0.434&0.426&0.514&\textcolor{blue}{0.572}&0.499&0.474&0.529&0.437&0.416&\textcolor{red}{0.606}\\
\hline
\end{tabular}
\end{center}
\label{tab:suc_attr}
\end{table*}

\begin{figure*}
	\begin{center}
		\includegraphics[width=0.497\linewidth]{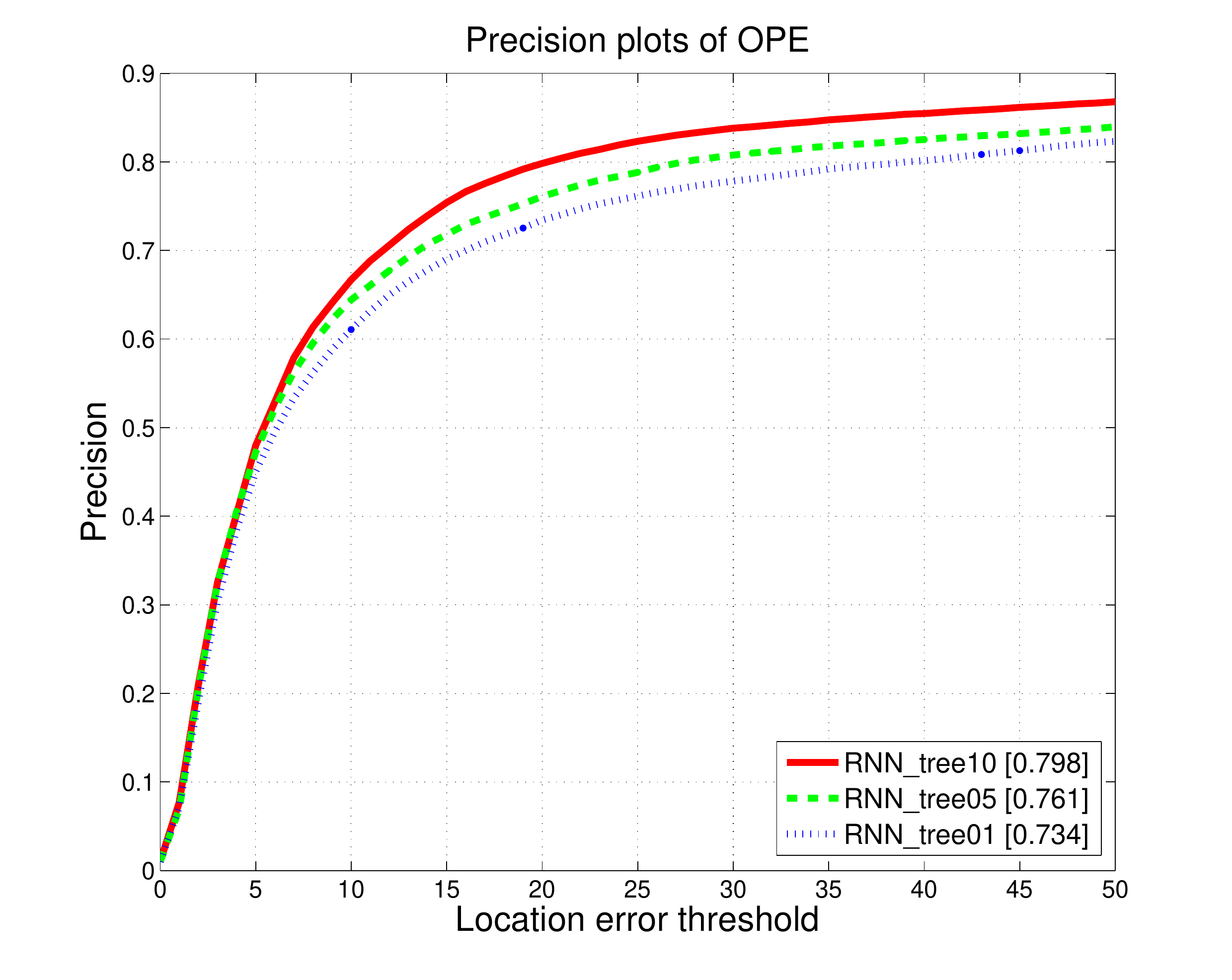}
		\includegraphics[width=0.497\linewidth]{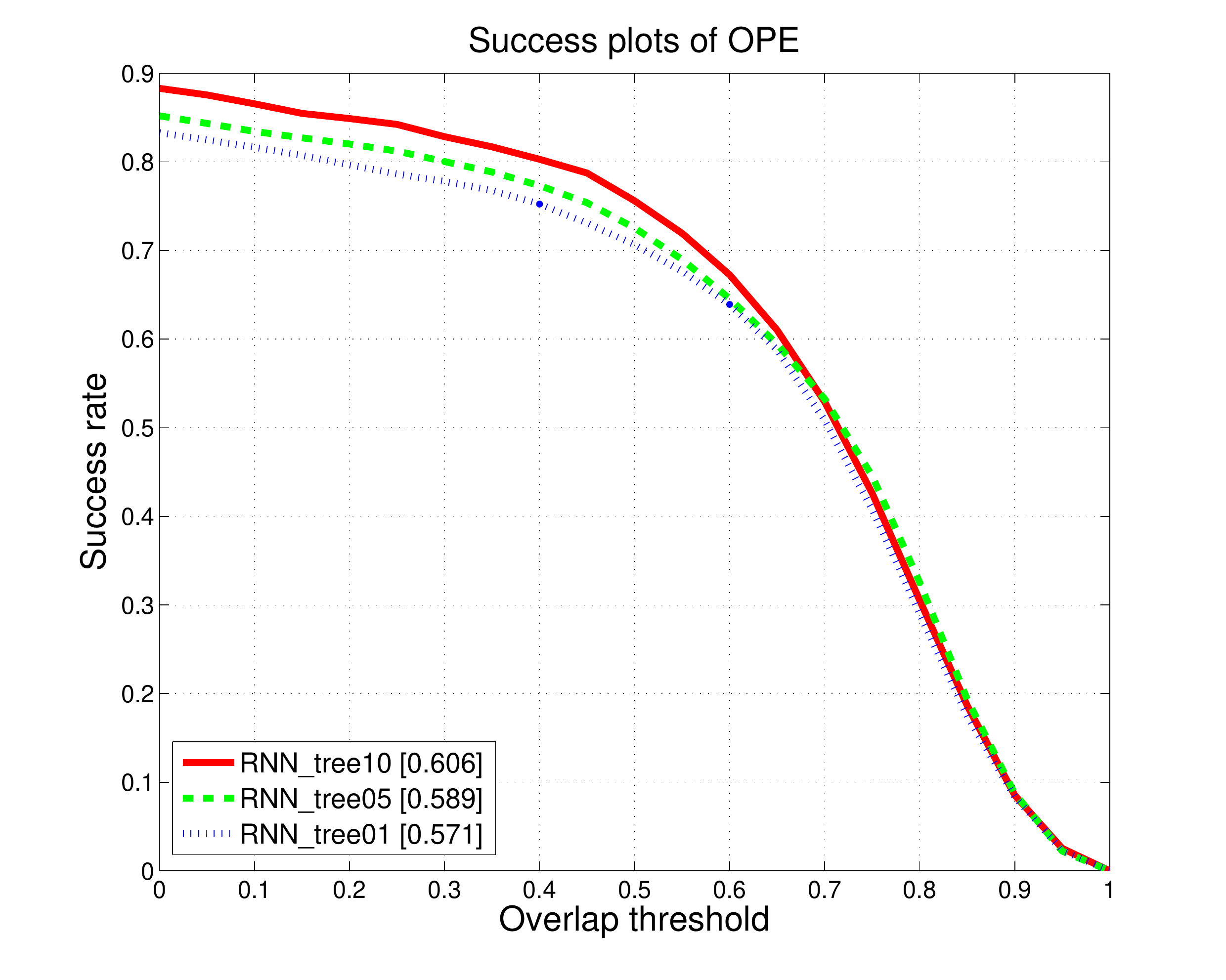}	
	\end{center}
	\caption{Precision and success plots of the tracking results from different variants of our tracker on all sequences of the benchmark \cite{DBLP:conf/cvpr/WuLY13}.}
	\label{fig:pre+suc_plot_baselineComp}
\end{figure*}

\textbf{Quantitative results.} The precision plot and the success plot of tracking results from the top $10$ trackers on all sequences of the benchmark \cite{DBLP:conf/cvpr/WuLY13} are presented in Figure~\ref{fig:pre+suc_plot_all}. We can find that our tracker outperforms the baseline ASLA \cite{DBLP:conf/cvpr/JiaLY12} in terms of both precision and success rate with large margins. We owe this significant improvement to our learned hierarchical features using tree structure based RNN which successfully encodes spatial information over local patches of target objects. Also, we find that our tracker using RNN features can achieve comparable performance against the state-of-the-art trackers.

To further evaluate tracking performance,  in Table~\ref{tab:pre_attr} and Table~\ref{tab:suc_attr}, we present the comparison results from the top $10$ trackers in terms of $11$ attributes mentioned in the benchmark \cite{DBLP:conf/cvpr/WuLY13}. We can find that our tracker using learned RNN features can consistently improve the baseline ASLA \cite{DBLP:conf/cvpr/JiaLY12} in terms of handling variational tracking challenges. Also, we can observe that our tracker achieves comparable performance against the state-of-the-art trackers in all attributes.

To investigate the effects of the numbers of random trees used in our RNN model, we present in Figure~\ref{fig:pre+suc_plot_baselineComp} the comparison results from different variants of our tracker in terms of both precision and success plots. We can find that increasing the number of random trees can significantly enhance tracking performance. It is because that learned hierarchical features based on more tree structures can encode more spatial information over local patches of target objects.

\begin{figure*}
	\begin{center}
		\includegraphics[width=0.18\linewidth, height=2.3cm]{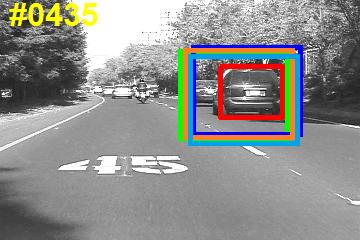}
		\includegraphics[width=0.18\linewidth, height=2.3cm]{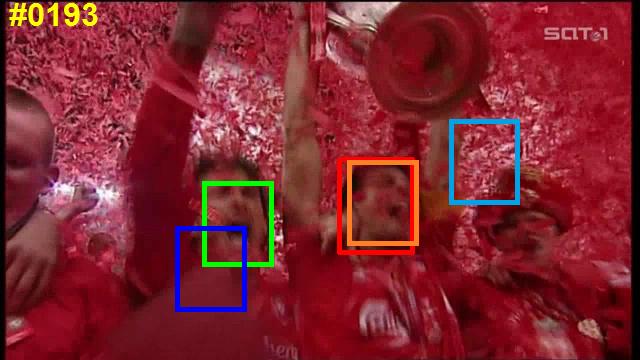}
		\includegraphics[width=0.18\linewidth, height=2.3cm]{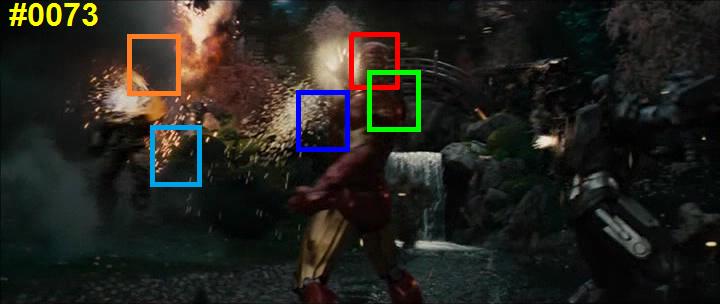}
		\includegraphics[width=0.18\linewidth, height=2.3cm]{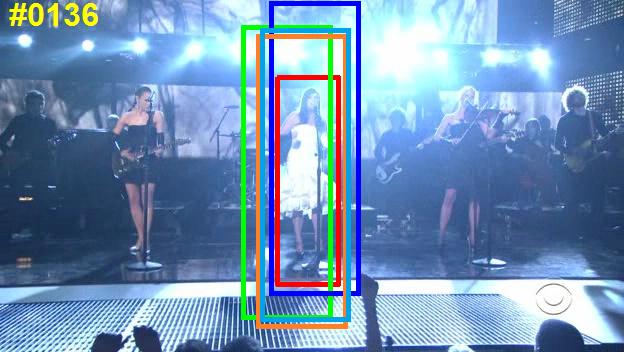}
		\includegraphics[width=0.18\linewidth, height=2.3cm]{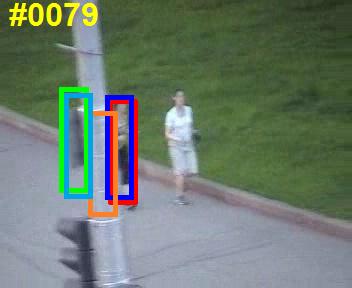}\\ \vspace{0.05in}
		
		\includegraphics[width=0.18\linewidth, height=2.3cm]{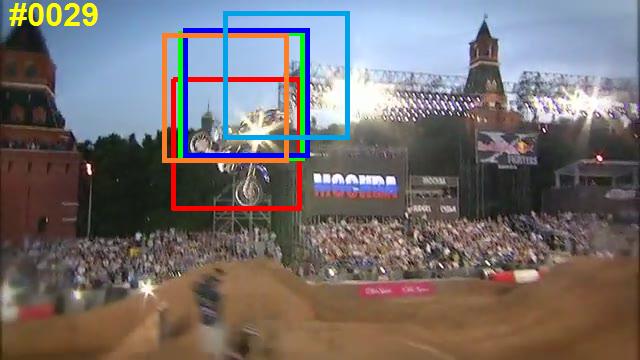}
		\includegraphics[width=0.18\linewidth, height=2.3cm]{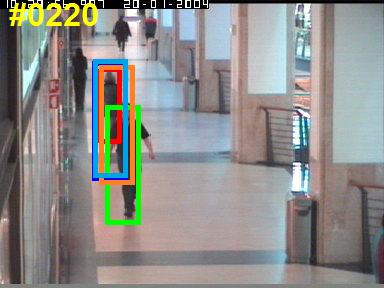}
		\includegraphics[width=0.18\linewidth, height=2.3cm]{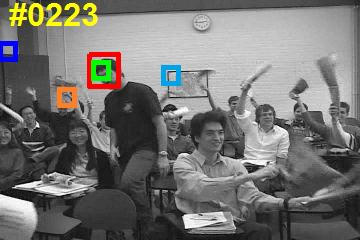}
		\includegraphics[width=0.18\linewidth, height=2.3cm]{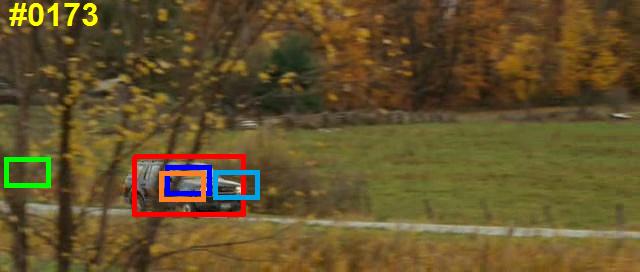}
		\includegraphics[width=0.18\linewidth, height=2.3cm]{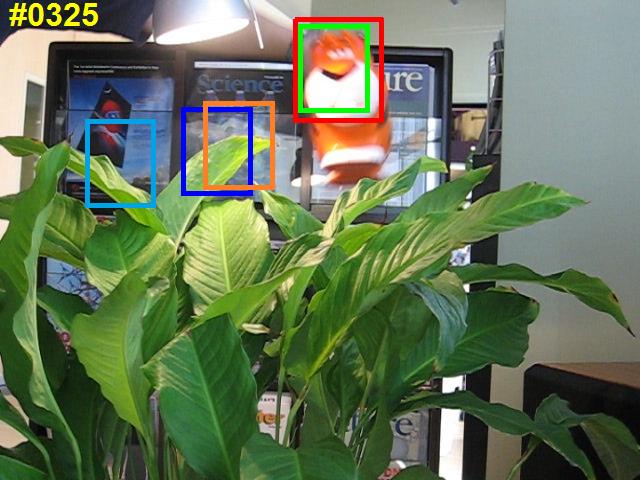}\\
			
	\end{center}
	\caption{Qualitative results comparing our tracker (red) with the other $4$ state-of-the-art trackers: MEEM \cite{DBLP:conf/eccv/ZhangMS14} (Green), TGPR \cite{DBLP:conf/eccv/GaoLHX14} (dark blue), KCF \cite{DBLP:journals/pami/HenriquesC0B15} (orange), Struck \cite{DBLP:conf/iccv/HareST11} (light blue) on $10$ sequences (car4, soccer, ironman, singer1, jogging-1, motorRolling, walking2, freeman4, carScale, tiger1).}
	\label{fig:qualitative}
\end{figure*}

\textbf{Qualitative results. } We present some qualitative results from MEEM \cite{DBLP:conf/eccv/ZhangMS14}, TGPR \cite{DBLP:conf/eccv/GaoLHX14}, KCF \cite{DBLP:journals/pami/HenriquesC0B15}, Struck \cite{DBLP:conf/iccv/HareST11} and our tracker using RNN features on $10$ sequences (car4, soccer, ironman, singer1, jogging-1, motorRolling, walking2, freeman4, carScale, tiger1) in Figure~\ref{fig:qualitative}, from which we can observe that our tracker can handle a large variety of tracking challenges, \eg scale variation, occlusion, motion blur, illumination variation and in-plane rotation. In particular, our tracker can achieve very promising performance in handling scale variation (car4, singer1, walking2, carScale). It is because that: i) the baseline tracker can generate candidates with different scales; ii) our learned hierarchical features encode spatial information over local patches of target regions; iii) RNN features are learned to discriminate target from background. The later two factors further enhance the first factor for handling scale changes. As a result, the bounding boxes of our tracking results can be accurately fit to target regions.

\section{Conclusions}
In this paper, we propose to learn hierarchical features for visual object tracking by using tree structure based RNN which can encode spatial information over local patches of target objects. For tree structures, we use a bottom-up greedy searching strategy to generate random trees. Given a test sequence, we learn RNN parameters to discriminate target from background in the first frame and then fix them for extracting RNN features from candidate regions in the subsequent frames. With learned RNN parameters, we create two dictionaries regarding target regions and corresponding local patches based on learned RNN features respectively from top and leaf nodes of multiple random trees in the first $10$ frames. In each of the subsequent frames, we conduct sparse dictionary coding on all candidate regions to find the optimal candidate as the new target location of the current frame. Additionally, we online update two dictionaries to handle target appearance changes. Experimental results demonstrate that our feature learning algorithm can significantly improve tracking performance.


\ifCLASSOPTIONcaptionsoff
  \newpage
\fi

\bibliographystyle{IEEEtran}
\bibliography{ref_v1}

\begin{thebibliography}{10}
\providecommand{\url}[1]{#1}
\csname url@samestyle\endcsname
\providecommand{\newblock}{\relax}
\providecommand{\bibinfo}[2]{#2}
\providecommand{\BIBentrySTDinterwordspacing}{\spaceskip=0pt\relax}
\providecommand{\BIBentryALTinterwordstretchfactor}{4}
\providecommand{\BIBentryALTinterwordspacing}{\spaceskip=\fontdimen2\font plus
\BIBentryALTinterwordstretchfactor\fontdimen3\font minus
  \fontdimen4\font\relax}
\providecommand{\BIBforeignlanguage}[2]{{%
\expandafter\ifx\csname l@#1\endcsname\relax
\typeout{** WARNING: IEEEtran.bst: No hyphenation pattern has been}%
\typeout{** loaded for the language `#1'. Using the pattern for}%
\typeout{** the default language instead.}%
\else
\language=\csname l@#1\endcsname
\fi
#2}}
\providecommand{\BIBdecl}{\relax}
\BIBdecl

\bibitem{DBLP:conf/cvpr/JiaLY12}
X.~Jia, H.~Lu, and M.~Yang, ``Visual tracking via adaptive structural local
  sparse appearance model,'' in \emph{{IEEE} Conference on Computer Vision and
  Pattern Recognition}, 2012, pp. 1822--1829.

\bibitem{DBLP:conf/nips/WangY13}
N.~Wang and D.~Yeung, ``Learning a deep compact image representation for visual
  tracking,'' in \emph{Advances in Neural Information Processing Systems},
  2013, pp. 809--817.

\bibitem{DBLP:conf/bmvc/LiLP14}
H.~Li, Y.~Li, and F.~Porikli, ``Deeptrack: Learning discriminative feature
  representations by convolutional neural networks for visual tracking,'' in
  \emph{British Machine Vision Conference}, 2014.

\bibitem{DBLP:journals/tip/WangLWCY15}
L.~Wang, T.~Liu, G.~Wang, K.~L. Chan, and Q.~Yang, ``Video tracking using
  learned hierarchical features,'' \emph{{IEEE} Transactions on Image
  Processing}, vol.~24, no.~4, pp. 1424--1435, 2015.

\bibitem{DBLP:conf/icml/HongYKH15}
S.~Hong, T.~You, S.~Kwak, and B.~Han, ``Online tracking by learning
  discriminative saliency map with convolutional neural network,'' in
  \emph{International Conference on Machine Learning}, 2015, pp. 597--606.

\bibitem{DBLP:conf/iccv/WangOWL15}
L.~Wang, W.~Ouyang, X.~Wang, and H.~Lu, ``Visual tracking with fully
  convolutional networks,'' in \emph{{IEEE} International Conference on
  Computer Vision}, 2015.

\bibitem{DBLP:conf/iccv/MaOHYY15}
C.~Ma, J.-B. Huang, X.~Yang, and M.-H. Yang, ``Hierarchical convolutional
  features for visual tracking,'' in \emph{{IEEE} International Conference on
  Computer Vision}, 2015.

\bibitem{DBLP:conf/cvpr/WuLY13}
Y.~Wu, J.~Lim, and M.~Yang, ``Online object tracking: {A} benchmark,'' in
  \emph{{IEEE} Conference on Computer Vision and Pattern Recognition}, 2013,
  pp. 2411--2418.

\bibitem{DBLP:journals/pami/SmeuldersCCCDS14}
A.~W.~M. Smeulders, D.~M. Chu, R.~Cucchiara, S.~Calderara, A.~Dehghan, and
  M.~Shah, ``Visual tracking: An experimental survey,'' \emph{{IEEE} Trans.
  Pattern Anal. Mach. Intell.}, vol.~36, no.~7, pp. 1442--1468, 2014.

\bibitem{DBLP:journals/ijcv/BlackJ98}
M.~J. Black and A.~D. Jepson, ``Eigentracking: Robust matching and tracking of
  articulated objects using a view-based representation,'' \emph{International
  Journal of Computer Vision}, vol.~26, no.~1, pp. 63--84, 1998.

\bibitem{DBLP:journals/ijcv/RossLLY08}
D.~A. Ross, J.~Lim, R.~Lin, and M.~Yang, ``Incremental learning for robust
  visual tracking,'' \emph{International Journal of Computer Vision}, vol.~77,
  no. 1-3, pp. 125--141, 2008.

\bibitem{DBLP:conf/iccv/MeiL09}
X.~Mei and H.~Ling, ``Robust visual tracking using $\ell_1$ minimization,'' in
  \emph{{IEEE} International Conference on Computer Vision}, 2009, pp.
  1436--1443.

\bibitem{DBLP:conf/cvpr/LiSS11}
H.~Li, C.~Shen, and Q.~Shi, ``Real-time visual tracking using compressive
  sensing,'' in \emph{{IEEE} Conference on Computer Vision and Pattern
  Recognition}, 2011, pp. 1305--1312.

\bibitem{DBLP:conf/cvpr/BaoWLJ12}
C.~Bao, Y.~Wu, H.~Ling, and H.~Ji, ``Real time robust {L1} tracker using
  accelerated proximal gradient approach,'' in \emph{{IEEE} Conference on
  Computer Vision and Pattern Recognition}, 2012, pp. 1830--1837.

\bibitem{DBLP:journals/ivc/McKennaRG99}
S.~J. McKenna, Y.~Raja, and S.~Gong, ``Tracking colour objects using adaptive
  mixture models,'' \emph{Image Vision Comput.}, vol.~17, no. 3-4, pp.
  225--231, 1999.

\bibitem{DBLP:journals/pami/ComaniciuRM03}
D.~Comaniciu, V.~Ramesh, and P.~Meer, ``Kernel-based object tracking,''
  \emph{{IEEE} Trans. Pattern Anal. Mach. Intell.}, vol.~25, no.~5, pp.
  564--575, 2003.

\bibitem{DBLP:conf/cvpr/KwonL10}
J.~Kwon and K.~M. Lee, ``Visual tracking decomposition,'' in \emph{{IEEE}
  Conference on Computer Vision and Pattern Recognition}, 2010, pp. 1269--1276.

\bibitem{DBLP:journals/pami/Avidan04}
S.~Avidan, ``Support vector tracking,'' \emph{{IEEE} Trans. Pattern Anal. Mach.
  Intell.}, vol.~26, no.~8, pp. 1064--1072, 2004.

\bibitem{DBLP:conf/iccv/TangBZT07}
F.~Tang, S.~Brennan, Q.~Zhao, and H.~Tao, ``Co-tracking using semi-supervised
  support vector machines,'' in \emph{{IEEE} International Conference on
  Computer Vision}, 2007, pp. 1--8.

\bibitem{DBLP:conf/iccv/HareST11}
S.~Hare, A.~Saffari, and P.~H.~S. Torr, ``Struck: Structured output tracking
  with kernels,'' in \emph{{IEEE} International Conference on Computer Vision},
  2011, pp. 263--270.

\bibitem{DBLP:conf/cvpr/BaiT12}
Y.~Bai and M.~Tang, ``Robust tracking via weakly supervised ranking {SVM},'' in
  \emph{{IEEE} Conference on Computer Vision and Pattern Recognition}, 2012,
  pp. 1854--1861.

\bibitem{DBLP:conf/cvpr/GrabnerB06}
H.~Grabner and H.~Bischof, ``On-line boosting and vision,'' in \emph{{IEEE}
  Computer Society Conference on Computer Vision and Pattern Recognition},
  2006, pp. 260--267.

\bibitem{DBLP:conf/iccv/LiuCL09}
R.~Liu, J.~Cheng, and H.~Lu, ``A robust boosting tracker with minimum error
  bound in a co-training framework,'' in \emph{{IEEE} International Conference
  on Computer Vision}, 2009, pp. 1459--1466.

\bibitem{DBLP:conf/cvpr/ZeislLSB10}
B.~Zeisl, C.~Leistner, A.~Saffari, and H.~Bischof, ``On-line semi-supervised
  multiple-instance boosting,'' in \emph{{IEEE} Conference on Computer Vision
  and Pattern Recognition}, 2010, p. 1879.

\bibitem{DBLP:conf/iccv/ZhangHML07}
X.~Zhang, W.~Hu, S.~J. Maybank, and X.~Li, ``Graph based discriminative
  learning for robust and efficient object tracking,'' in \emph{{IEEE}
  International Conference on Computer Vision}, 2007, pp. 1--8.

\bibitem{DBLP:conf/cvpr/BabenkoYB09}
B.~Babenko, M.~Yang, and S.~J. Belongie, ``Visual tracking with online multiple
  instance learning,'' in \emph{{IEEE} Computer Society Conference on Computer
  Vision and Pattern Recognition}, 2009, pp. 983--990.

\bibitem{DBLP:conf/eccv/WangHH10}
X.~Wang, G.~Hua, and T.~X. Han, ``Discriminative tracking by metric learning,''
  in \emph{European Conference on Computer Vision}, 2010, pp. 200--214.

\bibitem{DBLP:conf/cvpr/JiangLW12}
N.~Jiang, W.~Liu, and Y.~Wu, ``Order determination and sparsity-regularized
  metric learning adaptive visual tracking,'' in \emph{{IEEE} Conference on
  Computer Vision and Pattern Recognition}, 2012, pp. 1956--1963.

\bibitem{DBLP:conf/eccv/GaoLHX14}
J.~Gao, H.~Ling, W.~Hu, and J.~Xing, ``Transfer learning based visual tracking
  with gaussian processes regression,'' in \emph{European Conference on
  Computer Vision}, 2014, pp. 188--203.

\bibitem{DBLP:conf/emnlp/SocherPHNM11}
R.~Socher, J.~Pennington, E.~H. Huang, A.~Y. Ng, and C.~D. Manning,
  ``Semi-supervised recursive autoencoders for predicting sentiment
  distributions,'' in \emph{Proceedings of Conference on Empirical Methods in
  Natural Language Processing}, 2011, pp. 151--161.

\bibitem{DBLP:conf/emnlp/SocherHMN12}
R.~Socher, B.~Huval, C.~D. Manning, and A.~Y. Ng, ``Semantic compositionality
  through recursive matrix-vector spaces,'' in \emph{Proceedings of Joint
  Conference on Empirical Methods in Natural Language Processing and
  Computational Natural Language Learning}, 2012, pp. 1201--1211.

\bibitem{DBLP:conf/nips/SocherHPNM11}
R.~Socher, E.~H. Huang, J.~Pennington, A.~Y. Ng, and C.~D. Manning, ``Dynamic
  pooling and unfolding recursive autoencoders for paraphrase detection,'' in
  \emph{Advances in Neural Information Processing Systems}, 2011, pp. 801--809.

\bibitem{DBLP:conf/icml/SocherLNM11}
R.~Socher, C.~C. Lin, A.~Y. Ng, and C.~D. Manning, ``Parsing natural scenes and
  natural language with recursive neural networks,'' in \emph{Proceedings of
  International Conference on Machine Learning}, 2011, pp. 129--136.

\bibitem{DBLP:conf/nips/SocherHBMN12}
R.~Socher, B.~Huval, B.~P. Bath, C.~D. Manning, and A.~Y. Ng,
  ``Convolutional-recursive deep learning for 3d object classification,'' in
  \emph{Advances in Neural Information Processing Systems}, 2012, pp. 665--673.

\bibitem{goller1996learning}
C.~Goller and A.~Kuchler, ``Learning task-dependent distributed representations
  by backpropagation through structure,'' in \emph{IEEE International
  Conference on Neural Networks}, vol.~1, 1996, pp. 347--352.

\bibitem{nocedal1980updating}
J.~Nocedal, ``Updating quasi-newton matrices with limited storage,''
  \emph{Mathematics of Computation}, vol.~35, no. 151, pp. 773--782, 1980.

\bibitem{DBLP:conf/nips/KrizhevskySH12}
A.~Krizhevsky, I.~Sutskever, and G.~E. Hinton, ``Imagenet classification with
  deep convolutional neural networks,'' in \emph{Advances in Neural Information
  Processing Systems}, 2012, pp. 1106--1114.

\bibitem{DBLP:conf/cvpr/DinhVM11}
T.~B. Dinh, N.~Vo, and G.~G. Medioni, ``Context tracker: Exploring supporters
  and distracters in unconstrained environments,'' in \emph{{IEEE} Conference
  on Computer Vision and Pattern Recognition}, 2011, pp. 1177--1184.

\bibitem{DBLP:journals/pami/HenriquesC0B15}
J.~F. Henriques, R.~Caseiro, P.~Martins, and J.~Batista, ``High-speed tracking
  with kernelized correlation filters,'' \emph{{IEEE} Trans. Pattern Anal.
  Mach. Intell.}, vol.~37, no.~3, pp. 583--596, 2015.

\bibitem{DBLP:conf/eccv/ZhangMS14}
J.~Zhang, S.~Ma, and S.~Sclaroff, ``{MEEM:} robust tracking via multiple
  experts using entropy minimization,'' in \emph{European Conference on
  Computer Vision}, 2014, pp. 188--203.

\bibitem{DBLP:conf/cvpr/ZhongLY12}
W.~Zhong, H.~Lu, and M.~Yang, ``Robust object tracking via sparsity-based
  collaborative model,'' in \emph{{IEEE} Conference on Computer Vision and
  Pattern Recognition}, 2012, pp. 1838--1845.

\bibitem{DBLP:conf/cvpr/KalalMM10}
Z.~Kalal, J.~Matas, and K.~Mikolajczyk, ``{P-N} learning: Bootstrapping binary
  classifiers by structural constraints,'' in \emph{{IEEE} Conference on
  Computer Vision and Pattern Recognition}, 2010, pp. 49--56.

\bibitem{DBLP:conf/iccv/KwonL11}
J.~Kwon and K.~M. Lee, ``Tracking by sampling trackers,'' in \emph{{IEEE}
  International Conference on Computer Vision}, 2011, pp. 1195--1202.

\bibitem{DBLP:journals/tip/ZhongLY14}
W.~Zhong, H.~Lu, and M.~Yang, ``Robust object tracking via sparse collaborative
  appearance model,'' \emph{{IEEE} Transactions on Image Processing}, vol.~23,
  no.~5, pp. 2356--2368, 2014.

\end{thebibliography}

%
%

\end{document}